\def\year{2021}\relax
\documentclass[letterpaper]{article} 
\usepackage{aaai21}  
\usepackage{times}  
\usepackage{helvet} 
\usepackage{courier}  
\usepackage[hyphens]{url}  
\usepackage{graphicx} 
\usepackage{natbib}  
\usepackage{caption} 
\usepackage{latexsym}
\usepackage{amsmath}
\usepackage{amssymb}
\usepackage{microtype}
\usepackage{multirow}
\usepackage{booktabs}
\usepackage{marvosym}
\usepackage{bm}
\usepackage{threeparttable}
\usepackage{url}

\title{LET: Linguistic Knowledge Enhanced Graph Transformer \\for Chinese Short Text Matching}
\author{
    Boer Lyu, Lu Chen\thanks{The corresponding authors are Lu Chen and Kai Yu.}, Su Zhu, Kai Yu\footnotemark[1]\\
}
\affiliations{
    State Key Laboratory of Media Convergence Production Technology and Systems\\
    MoE Key Lab of Artificial Intelligence, AI Institute, Shanghai Jiao Tong University\\

    X-LANCE Lab, Department of Computer Science and Engineering\\
Shanghai Jiao Tong University, Shanghai, China\\
    \{boerlv, chenlusz, paul2204, kai.yu\}@sjtu.edu.cn

}

\begin{document}

\maketitle

\begin{abstract}
Chinese short text matching is a fundamental task in natural language processing. Existing approaches usually take Chinese characters or words as input tokens. They have two limitations: 1) Some Chinese words are polysemous, and semantic information is not fully utilized. 2) Some models suffer potential issues caused by word segmentation. Here we introduce HowNet as an external knowledge base and propose a Linguistic knowledge Enhanced graph Transformer (LET) to deal with word ambiguity. Additionally, we adopt the word lattice graph as input to maintain multi-granularity information. Our model is also complementary to pre-trained language models. Experimental results on two Chinese datasets show that our models outperform various typical text matching approaches. Ablation study also indicates that both semantic information and multi-granularity information are important for text matching modeling.
\end{abstract}

\section{Introduction}

Short text matching (STM) is generally regarded as a task of paraphrase identification or sentence semantic matching.
Given a pair of sentences, the goal of matching models is to predict their semantic similarity. 
It is widely used in question answer systems \cite{liu2018improved} and dialogue systems~\cite{gao2019neural,yu2014cognitive}.

Recent years have seen great progress in deep learning methods for text matching \cite{mueller2016siamese,gong2017natural,chen2017enhanced,lan2018neural}.
However, almost all of these models are initially proposed for English text matching. For Chinese language tasks, early work utilizes Chinese characters as input to the model, or first segment each sentence into words, and then take these words as input tokens. Although character-based models can overcome the problem of data sparsity to some degree~\cite{li2019word}, the main drawback is that explicit word information is not fully utilized, which has been demonstrated to be useful for semantic matching \cite{li2019enhancing}. 

\begin{figure}[t]
    \centering
    \includegraphics[width=0.95\columnwidth]{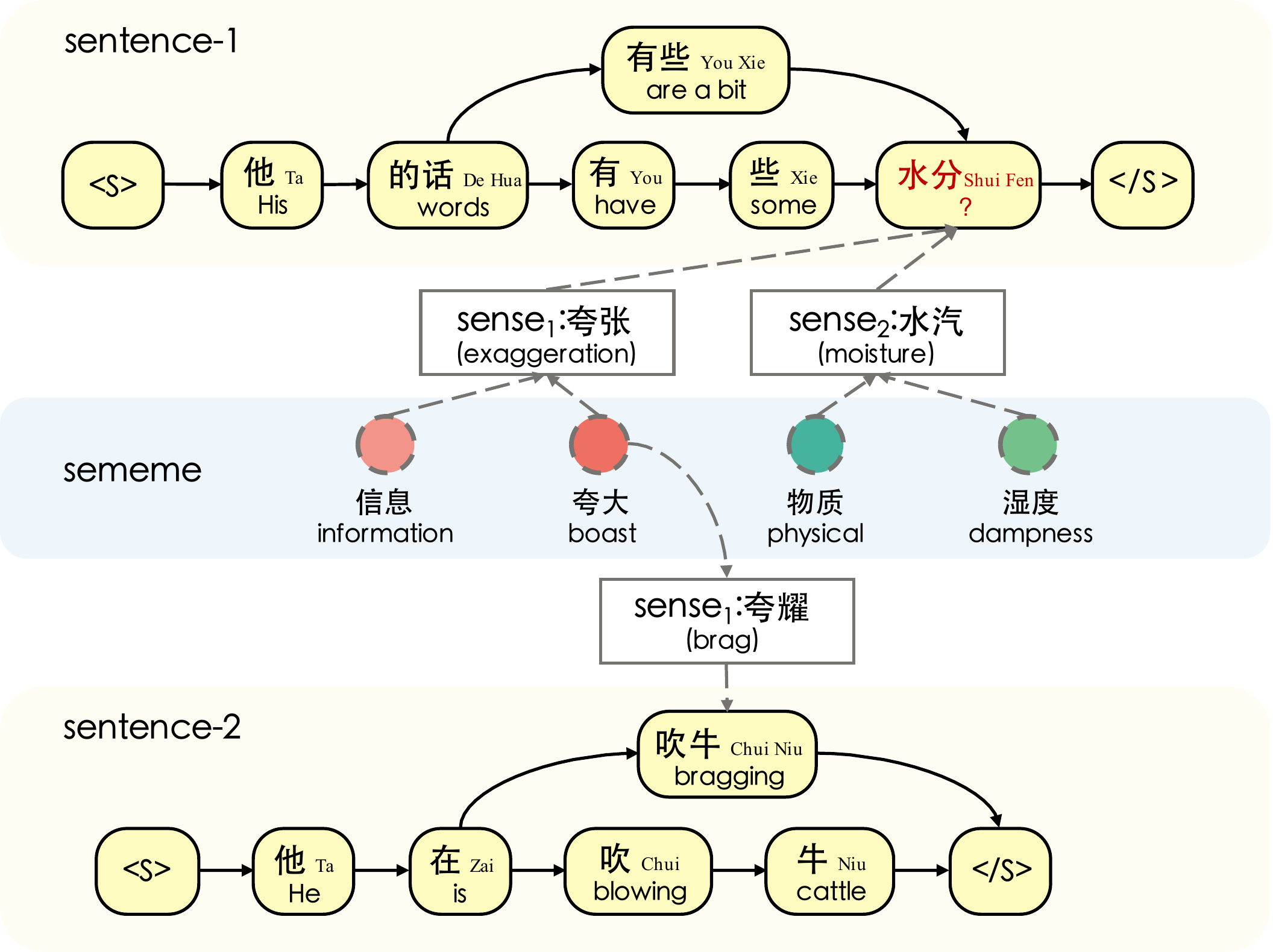}
    \caption{An example of word segmentation and the potential word ambiguity.}
    \label{fig:example}
\end{figure}

However, a large number of Chinese words are polysemous, which brings great difficulties to semantic understanding~\cite{xu2016improve}. 
Word polysemy in short text is more an issue than that in long text because short text usually has less contextual information, so it is extremely hard for models to capture the correct meaning. As shown in Fig. \ref{fig:example}, the word in red in sentence-1 actually has two meanings: one is to describe bragging (\emph{exaggeration}) and another is \emph{moisture}. 
Intuitively, if other words in the context have similar or related meanings, the probability of them will increase. 
To integrate semantic information of words, we introduce HowNet~\cite{dong2003hownet} as an external knowledge base. In the view of HowNet, words may have multiple senses/meanings and each sense has several sememes to represent it. For instance, the first sense \emph{exaggeration} indicates some boast information in his words. Therefore, it has sememes {\tt information} and {\tt boast}. Similarly, we can also find the sememe {\tt boast} describing the sense \emph{brag} which belongs to the word ``ChuiNiu (bragging)'' in sentence-2. In this way, model can better determine the sense of words and perceive that two sentences probably have the same meaning.

Furthermore, word-based models often encounter some potential issues caused by word segmentation. If the word segmentation fails to output ``ChuiNiu (bragging)'' in sentence-2, we will lose useful sense information. In Chinese, ``Chui (blowing)'' ``Niu (cattle)'' is a bad segmentation, which deviates the correct meaning of ``ChuiNiu (bragging)''.
To tackle this problem, many researchers propose word lattice graphs~\cite{lai2019lattice,li2020flat,chen2020neural}, where they retain words existing in the word bank so that various segmentation paths are kept. It has been shown that multi-granularity information is important for text matching. 

In this paper, we propose a Linguistic knowledge Enhanced graph Transformer (LET) to consider both semantic information and multi-granularity information. LET takes a pair of word lattice graphs as input. Since keeping all possible words will introduce a lot of noise, we use several segmentation paths to form our lattice graph and construct a set of senses according to the word. Based on HowNet, each sense has several sememes to represent it. In the input module, starting from the pre-trained sememe embeddings provided by OpenHowNet~\cite{qi2019openhownet}, we obtain the initial sense representation using a multi-dimensional graph attention transformer (MD-GAT, see Sec. 3.1). Also, we get the initial word representation by aggregating features from the character-level transformer encoder using an Att-Pooling (see Sec. 4.1). Then it is followed by SaGT layers (see Sec. 4.2), which fuse the information between words and semantics. In each layer, we first update the nodes' sense representation and then updates word representation using MD-GAT. As for the sentence matching layer (see Sec. 4.3), we convert word representation to character level and share the message between texts. Moreover, LET can be combined with pre-trained language models, e.g. BERT~\cite{devlin2019bert}. It can be regarded as a method to integrate word and sense information into pre-trained language models during the fine-tuning phase. 

 Contributions in this work are summarized as: a) We propose a novel enhanced graph transformer using linguistic knowledge to moderate word ambiguity. b) Empirical study on two Chinese datasets shows that our model outperforms not only typical text matching models but also the pre-trained model BERT as well as some variants of BERT. c) We demonstrate that both semantic information and multi-granularity information are important for text matching modeling, especially on shorter texts.

\section{Related Work}

{\bf Deep Text Matching Models} based on 
deep learning have been widely adopted for short text matching. They can fall into two categories: representation-based methods~\cite{he2016text,lai2019lattice} and interaction-based methods~\cite{wang2017bilateral,chen2017enhanced}. Most representation-based methods are based on Siamese architecture, which has two symmetrical networks (e.g. LSTMs and CNNs) to extract high-level features from two sentences. Then, these features are compared to predict text similarity. Interaction-based models incorporate interactions features between all word pairs in two sentences. They generally perform better than representation-based methods. Our proposed method belongs to interaction-based methods.

{\bf Pre-trained Language Models}, e.g. BERT, have shown its powerful performance on various natural language processing (NLP) tasks including text matching. For Chinese text matching, BERT takes a pair of sentences as input and each Chinese character is a separated input token. It has ignored word information. To tackle this problem, some Chinese variants of original BERT have been proposed, e.g. BERT-wwm~\cite{cui2019pre} and ERNIE~\cite{sun2019ernie}. They take the word information into consideration based on the whole word masking mechanism during pre-training. However, 
the pre-training process of a word-considered BERT requires a lot of time and resources. Thus, Our model takes pre-trained language model as initialization and utilizes word information to fine-tune the model.

\section{Background}

In this section, we introduce graph attention networks (GATs) and HowNet, which are the basis of our proposed models in the next section.
\subsection{Graph Attention Networks}



Graph neural networks (GNNs)~\cite{scarselli2008graph} are widely applied in various NLP tasks, such as text classifcation~\cite{yao2019graph}, text generation~\cite{zhao2020line}, dialogue policy optimization~\cite{ chen2018structured,chen2018policy,chen2019agentgraph,chen2020distributed} and dialogue state tracking~\cite{chen2020schema,zhu2020efficient}, etc.
GAT is a special type of GNN that operates on graph-structured data with attention mechanisms. Given a graph $G = (\mathcal{V}, \mathcal{E})$, where $\mathcal{V}$ and $\mathcal{E}$ are the set of nodes $x_i$ and the set of edges, respectively. $\mathcal{N}^{+}(x_i)$ is the set including the node $x_i$ itself and the nodes which are directly connected by $x_i$. 

Each node $x_i$ in the graph has an initial feature vector $\mathbf{h}_i^0 \in \mathbb{R}^d$, where
$d$ is the feature dimension. The representation of each node is iteratively updated by the graph attention operation. At the $l$-th step, each node $x_i$ aggregates context information by attending over its neighbors and itself. The updated representation $\mathbf{h}^l_i$ is calculated by the weighted average of the connected nodes,
\begin{equation}
\begin{split}
    \mathbf{h}_i^l 
    = \sigma\left(\sum_{x_j \in \mathcal{N}^{+}(x_i)} \alpha_{ij}^l \cdot \left( \mathbf{W}^l \mathbf{h}_j^{l-1}\right)\right),
\end{split}
\label{eq:selfatt}
\end{equation}
where $\mathbf{W}^l \in \mathbb{R}^{d \times d}$ is a learnable parameter, and
$\sigma(\cdot)$ is a nonlinear activation function, e.g. ReLU. The attention coefficient $\alpha_{ij}^l$ is the normalized similarity of the embedding between the two nodes $x_i$ and $x_j$ in a unified space, i.e. 
\begin{equation}
\begin{split}
    \alpha_{ij}^l &= \mbox{softmax}_{j} \ f^l_{sim}\left( \mathbf{h}^{l-1}_i, \  \mathbf{h}^{l-1}_j \right) \\
    & = \mbox{softmax}_{j}\left(\mathbf{W}^l_{q} \mathbf{h}_i^{l-1}\right)^T\left(\mathbf{W}^l_{k} \mathbf{h}_j^{l-1}\right),
    \label{eq:attention}
\end{split}
\end{equation}
where $\mathbf{W}^l_{q}$ and $\mathbf{W}^l_{k} \in \mathbb{R}^{d \times d}$ are learnable parameters for projections.

Note that, in Eq. (\ref{eq:attention}), $\alpha_{ij}^l$ is a scalar, which means that all dimensions in $\mathbf{h}_j^{l-1}$ are treated equally. This may limit the capacity to model complex dependencies. 
Following \citeauthor{shen2018disan}~\shortcite{shen2018disan}, we replace the vanilla attention with multi-dimensional attention.
Instead of computing a single scalar score, for each embedding $\mathbf{h}_j^{l-1}$, it first computes a feature-wise score vector, and then normalizes 
it with feature-wised multi-dimensional softmax (MD-softmax),
\begin{equation}
\begin{split}
    \bm{\alpha}_{ij}^l 
    = &\mbox{MD-softmax}_j\left( \hat{\alpha}^l_{ij} + f^l_{m}\left(\mathbf{h}_j^{l-1}\right)\right),
\end{split}
\end{equation}
where $\hat{\alpha}_{ij}^l$ is a scalar calculated by the similarity function $f^l_{sim} ( \cdot)$ in Eq. (\ref{eq:attention}), and $f^l_{m}(\cdot)$ is a vector. The addition in above equation means the scalar will be added to every element of the vector. $\hat{\alpha}_{ij}^l$ is utilized to model the pair-wised dependency of two nodes, while $f^l_{m}(\cdot)$ is used to estimate the contribution of each feature dimension of $\mathbf{h}_j^{l-1}$,
\begin{equation}
    f^l_{m}(\mathbf{h}_j^{l-1}) =  \mathbf{W}^l_{2}  \sigma \left(\mathbf{W}^l_{1} \mathbf{h}_j^{l-1} + \mathbf{b}^l_{1} \right) + \mathbf{b}^l_{2},
\end{equation}
where $\mathbf{W}_{1}^l$, $\mathbf{W}_{2}^l$, $\mathbf{b}^l_{1}$ and $\mathbf{b}^l_{2}$ are learnable parameters. 
With the score vector $\bm{\alpha}_{ij}^l$, Eq. (\ref{eq:selfatt}) will be accordingly revised as
\begin{equation}
\begin{split}
    \mathbf{h}_i^l 
    &= \sigma\left(\sum_{x_j \in \mathcal{N}^{+}(x_i)} \bm{\alpha}_{ij}^l \odot \left( \mathbf{W}^l \mathbf{h}_j^{l-1} \right) \right),
\end{split}
\label{eq:mdatt}
\end{equation}
where $\odot$ represents element-wise product of two vectors.
For brevity, we use $\mbox{MD-GAT}(\cdot)$ to denote the updating process using multi-dimensional attention mechanism, and rewrite Eq. (\ref{eq:mdatt}) as follows,
\begin{equation}
\begin{split}
    \mathbf{h}_i^l = \mbox{MD-GAT}\left(\mathbf{h}_i^{l-1}, \ \left\{ \mathbf{h}^{l-1}_j | x_j \in \mathcal{N}^{+}(x_i) \right\} \right). \\
\end{split}
\label{eq:mdatt_r}
\end{equation}

After $L$ steps of updating, each node will finally have a context-aware representation $\mathbf{h}_i^L$. In order to achieve a stable training process, we also employ a residual connection followed by a layer normalization between two graph attention layers.

\subsection{HowNet}

\begin{figure}[h]
    \centering
    \includegraphics[width=0.9\columnwidth]{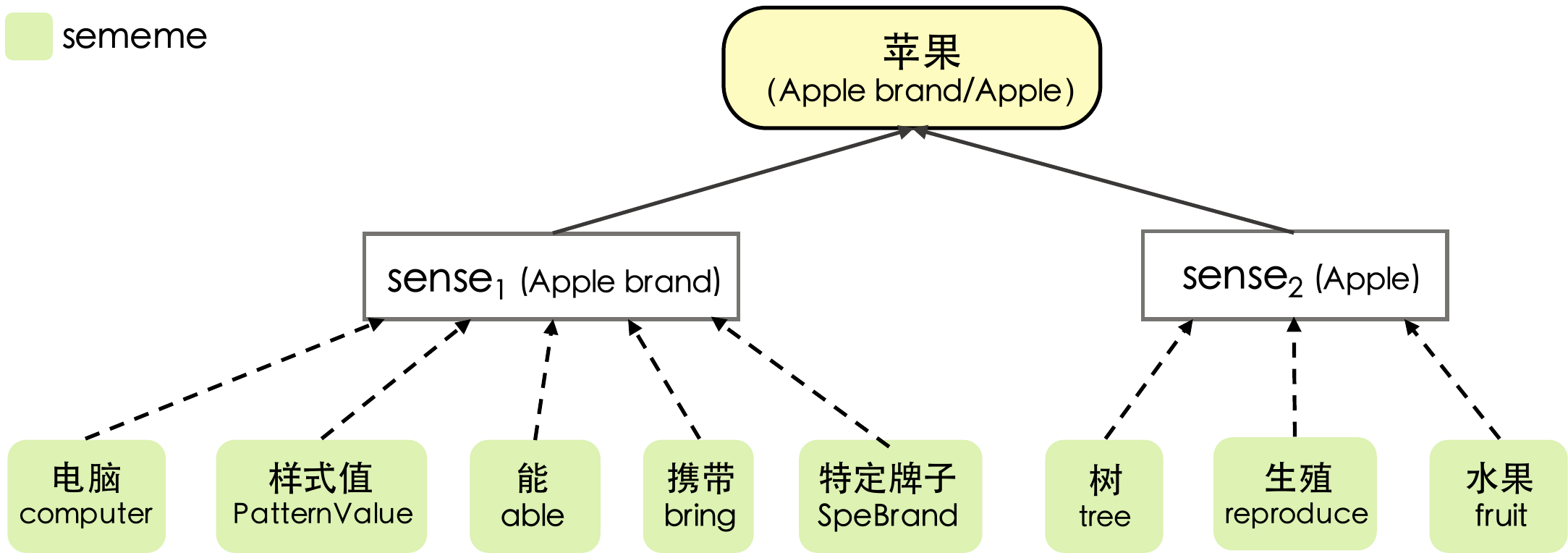}
    \caption{An example of the HowNet structure.}
    \label{fig:sense}
\end{figure}

HowNet~\cite{dong2003hownet} is an external knowledge base that manually annotates each Chinese word sense with one or more relevant sememes. The philosophy of HowNet regards sememe as an atomic semantic unit. Different from WordNet~\cite{miller1995wordnet}, it emphasizes that the parts and attributes of a concept can be well represented by sememes. HowNet has been widely utilized in many NLP tasks such as word similarity computation~\cite{liu2002word}, sentiment analysis~\cite{xianghua2013multi}, word representation learning~\cite{niu2017improved} and language modeling~\cite{gu2018language}.

An example is illustrated in Fig. \ref{fig:sense}. 
The word ``Apple'' has two senses including \emph{Apple Brand} and \emph{Apple}. The sense \emph{Apple Brand} has five sememes including {\tt computer}, {\tt PatternValue}, {\tt able}, {\tt bring} and {\tt SpecificBrand}, which describe the exact meaning of sense.

\section{Proposed Approach}
\label{sec:let}

\begin{figure}[h]
    \centering
    \includegraphics[width=1\columnwidth]{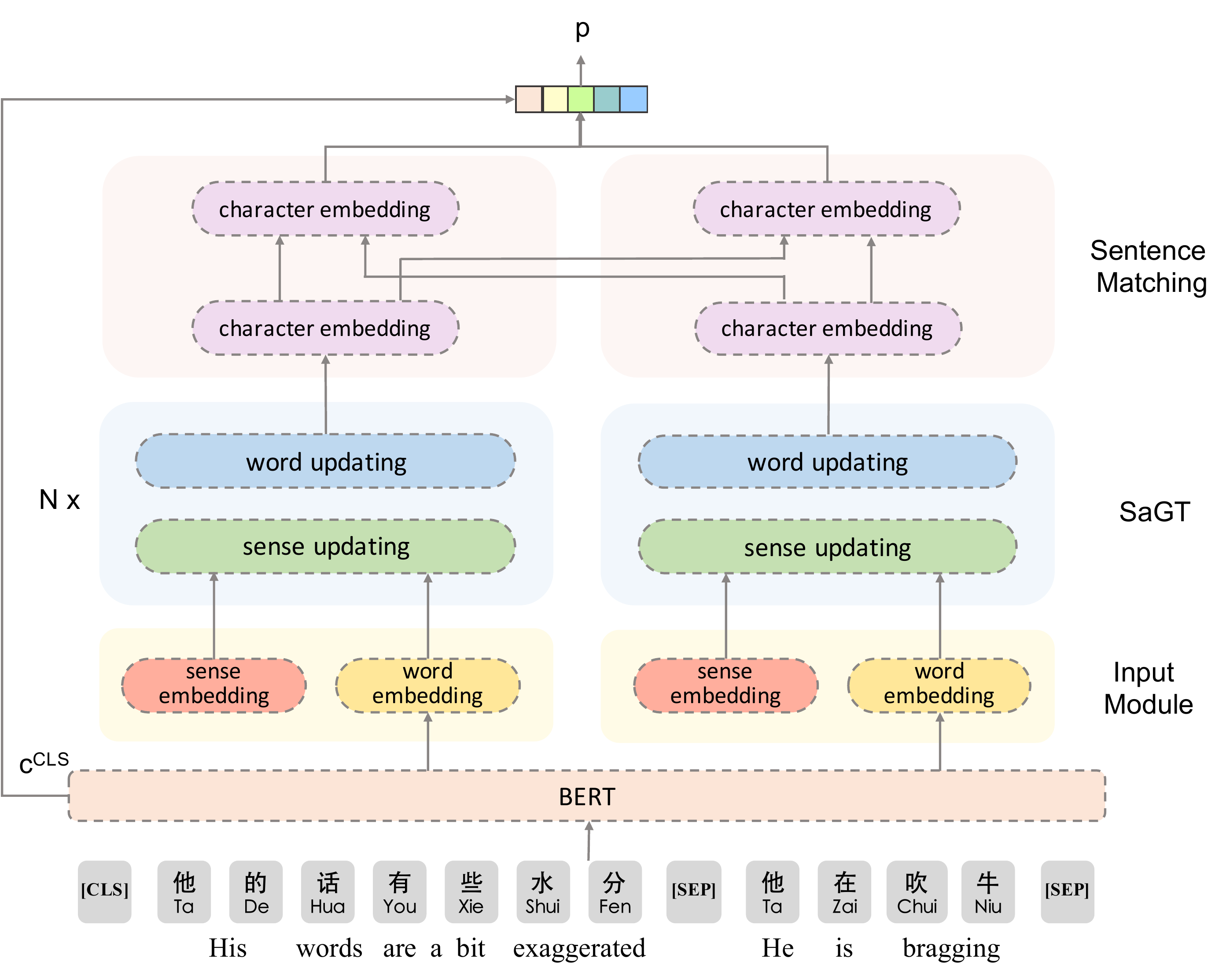}
    \caption{The framework of our proposed LET model.}
    \label{fig:net}
\end{figure}

First, we define the Chinese short text matching task in a formal way.
Given two Chinese sentences $\mathcal{C}^{a}=\{c_1^a, c_2^a, \cdots, c_{T_a}^a\}$ and $\mathcal{C}^{b}=\{c_1^b, c_2^b, \cdots, c_{T_b}^b\}$, the goal of a text matching model $f(\mathcal{C}^{a},\mathcal{C}^{b})$ is to predict whether the semantic meaning of $\mathcal{C}^{a}$ and $\mathcal{C}^{b}$ is equal. Here, $c_t^a$ and $c_{t'}^b$ represent the $t$-th and $t'$-th Chinese character in two sentences respectively, and $T_a$ and $T_b$ denote the number of characters in the sentences. 

\begin{figure}[t]
    \centering
    \includegraphics[width=1\columnwidth]{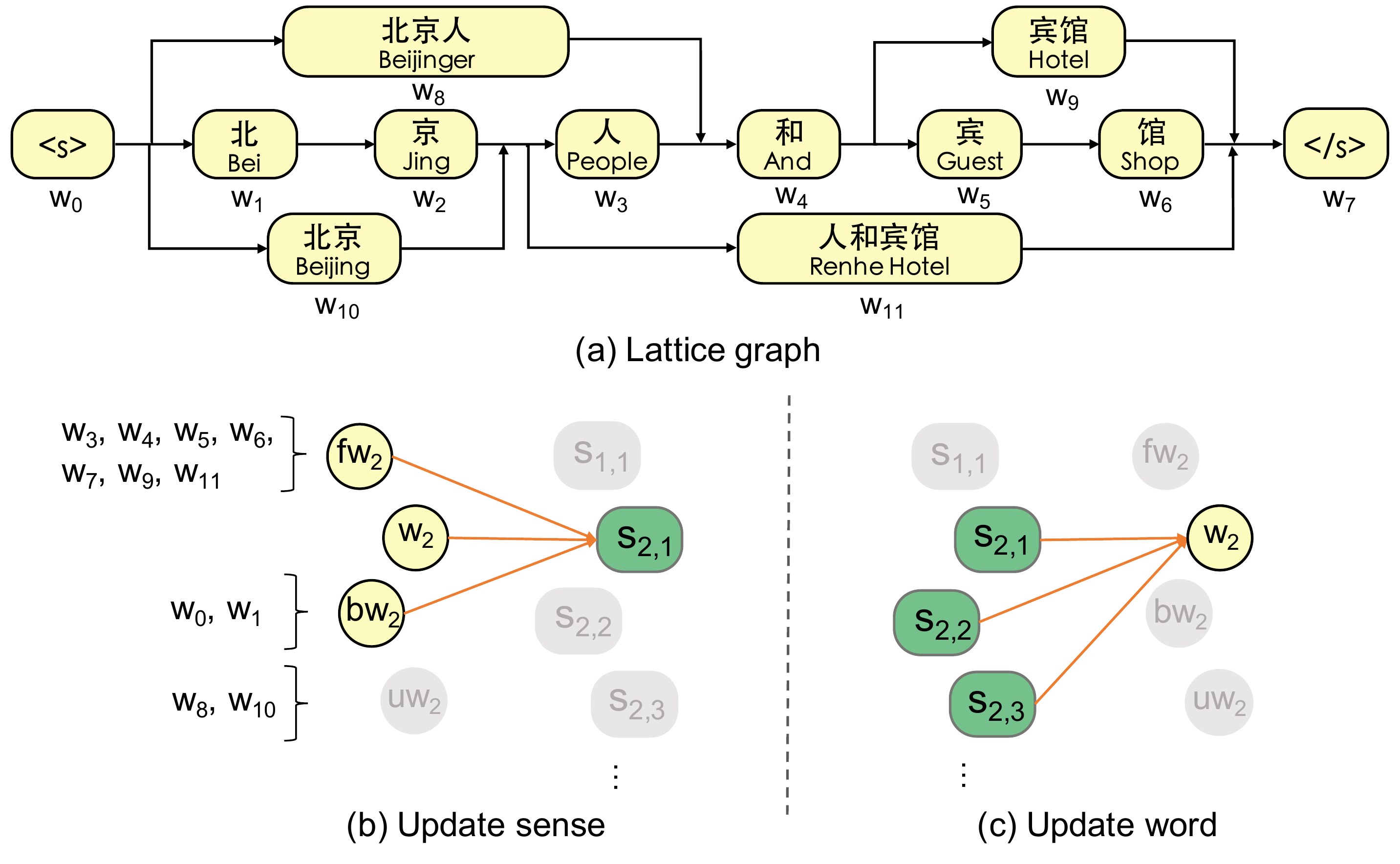}
    \caption{(a) is an example of lattice graph. (b) shows the process of sense updating. $\text{fw}_2$ and $\text{bw}_2$ refer to the words in forward and backward directions of $\text{w}_2$ respectively. $\text{uw}_2$ is the words that $\text{w}_2$ cannot reach. (c) is word updating; we will not update the corresponding word representation if the word is not in HowNet.}
    \label{fig:update}
\end{figure}

In this paper, we propose a linguistic knowledge enhanced matching model. Instead of segmenting each sentence into a word sequence, we use three segmentation tools and keep these segmentation paths to form a word lattice graph $G=(\mathcal{V},\mathcal{E})$ (see Fig. \ref{fig:update} (a)). $\mathcal{V}$ is the set of nodes and $\mathcal{E}$ is the set of edges.
Each node $x_i \in \mathcal{V}$ corresponds to a word $w_i$ which is a character subsequence starting from the $t_1$-th character to the $t_2$-th character in the sentence. As introduced in Sec. 1, we can obtain all senses of a word $w_i$ by retrieving the HowNet.

For two nodes $x_i \in \mathcal{V}$ and $x_j \in \mathcal{V}$,
if $x_i$ is adjacent to $x_j$ in the original sentence, then there is an edge between them.
$\mathcal{N}_{fw}^{+}(x_i)$ is the set including $x_i$ itself and all its reachable nodes in its forward direction, while $\mathcal{N}_{bw}^{+}(x_i)$ is the set including $x_i$ itself and all its reachable nodes in its backward direction.

Thus for each sample, we have two graphs $G^a=(\mathcal{V}^a,\mathcal{E}^a)$ and $G^b=(\mathcal{V}^b,\mathcal{E}^b)$, and our graph matching model is to predict their similarity.
As shown in Fig. \ref{fig:net}, LET consists of four components: input module, semantic-aware graph transformer (SaGT), sentence matching layer and relation classifier. The input module outputs the initial contextual representation for each word $w_i$ and the initial semantic representation for each sense. The semantic-aware graph transformer iteratively updates the word representation and sense representation, and fuses useful information from each other. The sentence matching layer first incorporates word representation into character level, and then matches two character sequences with the bilateral multi-perspective matching mechanism. The relation classifier takes the sentence vectors as input and predicts the relation of two sentences.


\subsection{Input Module}

\subsubsection{Contextual Word Embedding}
\label{sec:cwe}


For each node $x_i$ in graphs, the initial representation of word $w_i$ is the attentive pooling of contextual character representations. 
Concretely, we first concat the original character-level sentences to form a new sequence $\mathcal{C} = \{[\text{CLS}], c_1^a, \cdots, c_{T_a}^a, [\text{SEP}], c_1^b, \cdots, c_{T_b}^b, [\text{SEP}]\}$, and then feed them to the BERT model to obtain the contextual representations for each character $\{\mathbf{c}^{\text{CLS}}, \mathbf{c}_1^a$, $\cdots$, $\mathbf{c}_{T_a}^a$, $\mathbf{c}^{\text{SEP}}$, $\mathbf{c}_1^b$, $\cdots$, $\mathbf{c}_{T_b}^b$, $\mathbf{c}^{\text{SEP}}\}$. Assuming that the word $w_i$ consists of some consecutive character tokens $\{c_{t_1}, c_{{t_1}+1}, \cdots, c_{t_2}\}$\footnote{For brevity, the superscript of $c_{k}\ (t_1 \leq k \leq t_2)$ is omitted.}, a feature-wised score vector is calculated with a feed forward network (FFN) with two layers for each character $c_{k} \ (t_1 \leq k \leq t_2)$, and then normalized with a feature-wised  multi-dimensional softmax (MD-softmax),
\begin{equation}
   \mathbf{u}_{k}=\mbox{MD-softmax}_{k}\left( \text{FFN}(\mathbf{c}_{k}) \right),
 \label{eq:attpool-1}
\end{equation}
The corresponding character embedding $\mathbf{c}_{k}$ is weighted with the normalized scores $\mathbf{u}_{k}$ to obtain the contextual word embedding,
\begin{equation}
   \mathbf{v}_i = \sum_{k=t_1}^{t_2} \mathbf{u}_{k} \odot  \mathbf{c}_{k},
\label{eq:attpool-2}
\end{equation}
For brevity, we use $\mbox{Att-Pooling}(\cdot)$ to rewrite Eq. (\ref{eq:attpool-1}) and Eq. (\ref{eq:attpool-2}) for short, i.e. 
\begin{equation}
\mathbf{v}_i=\mbox{Att-Pooling}\left(\{\mathbf{c}_k | t_1 \leq k \leq t_2\}\right).
\end{equation}


\subsubsection{Sense Embedding}


The word embedding $\mathbf{v}_i$ described in Sec. \ref{sec:cwe} contains only contextual character information, which may suffer from the issue of polysemy in Chinese. In this paper, we incorporate HowNet as an external knowledge base to express the semantic information of words. 


For each word $w_i$, we denote the set of senses as $\mathcal{S}^{(w_i)}=\{s_{i,1}, s_{i,2}, \cdots, s_{i,K}\}$. $s_{i,k}$ is the $k$-th sense of $w_i$ and we denote its corresponding sememes as $\mathcal{O}^{(s_{i,k})}=\{o_{i,k}^1, o_{i,k}^2, \cdots, o_{i,k}^M\}$.
In order to get the embedding $\mathbf{s}_{i,k}$ for each sense $s_{i,k}$, we first obtain the representation $\mathbf{o}_{i,k}^m$ for each sememe $o_{i,k}^m$ with  multi-dimensional attention function,
\begin{equation}
\begin{split}
    \mathbf{o}^{m}_{i,k} &= \mbox{MD-GAT}\left(\mathbf{e}_{i,k}^{m}, \ \left\{\mathbf{e}_{i,k}^{m'} | o_{i,k}^{m'} \in \mathcal{O}^{(s_{i,k})} \right\} \right), \\
\end{split}
\end{equation}
where $\mathbf{e}_{i,k}^{m}$ is the embedding vector for sememe $o_{i,k}^m$ produced through
the Sememe Attention over Target model (SAT) \cite{niu2017improved}. Then, for each sense $s_{i,k}$, its embedding $\mathbf{s}_{i,k}$ is obtained with attentive pooling of all sememe representations,
\begin{equation}
    \mathbf{s}_{i,k} = \mbox{Att-Pooling}\left(\left\{ \mathbf{o}^{m}_{i,k}  |  o_{i,k}^{m} \in \mathcal{O}^{(s_{i,k})} \right\} \right).
\end{equation}



\subsection{Semantic-aware Graph Transformer}

For each node $x_i$ in the graph, the word embedding $\mathbf{v}_i$ only contains the contextual information while the sense embedding $\mathbf{s}_{i,k}$ only contains linguistic knowledge. In order to harvest useful information from each other, we propose a semantic-aware graph transformer (SaGT). It first takes $\mathbf{v}_i$ and $\mathbf{s}_{i,k}$ as initial word representation $\mathbf{h}^0_i$ for word $w_i$ and initial sense representation $\mathbf{g}^0_{i,k}$ for sense $s_{i,k}$ respectively, and then iteratively updates them with two sub-steps.

\subsubsection{Updating Sense Representation} At $l$-th iteration, the first sub-step is to update sense representation from $\mathbf{g}^{l-1}_{i,k}$ to $\mathbf{g}^l_{i,k}$. For a word with multiple senses, which sense should be used is usually determined by the context in the sentence. Therefore, when updating the representation, each sense will first aggregate useful information from words in forward and backward directions of $x_i$,
\begin{equation}
\begin{split}
    \mathbf{m}^{l,fw}_{i,k} &= \mbox{MD-GAT}\left(\mathbf{g}_{i,k}^{l-1}, \ \left\{\mathbf{h}_j^{l-1} | x_j \in \mathcal{N}_{fw}^{+}(x_i) \right\} \right), \\
    \mathbf{m}^{l,bw}_{i,k} &= \mbox{MD-GAT}\left(\mathbf{g}_{i,k}^{l-1}, \ \left\{\mathbf{h}_j^{l-1} | x_j \in \mathcal{N}_{bw}^{+}(x_i) \right\} \right), \\
\end{split}
\end{equation}
where two multi-dimensional attention functions $\mbox{MD-GAT}(\cdot)$ have different parameters. Based on $\mathbf{m}^l_{i,k}=[\mathbf{m}^{l,fw}_{i,k}, \mathbf{m}^{l,bw}_{i,k}]$~\footnote{$[\cdot,\cdot]$ denotes the concatenation of vectors.}, each sense updates its representation with a gate recurrent unit (GRU) \cite{cho2014learning},
\begin{equation}
    \mathbf{g}^l_{i,k} = \mbox{GRU}\left( \mathbf{g}_{i,k}^{l-1}, \ \mathbf{m}^l_{i,k}\right).
\label{eq:gru_sense}
\end{equation}

It is notable that we don't directly use $\mathbf{m}_{i,k}^l$ as the new representation $\mathbf{g}_{i,k}^{l}$ of sense $s_{i,k}$. 
The reason is that $\mathbf{m}_{i,k}^l$ only contains contextual information, and we need to utilize a gate, e.g. GRU, to control the fusion of contextual information and semantic information.

\subsubsection{Updating Word Representation} The second sub-step is to update the word representation from $\mathbf{h}^{l-1}_{i}$ to $\mathbf{h}^{l}_{i}$ based on the updated sense representations $\mathbf{g}_{i,k}^l \ (1\leq k \leq K)$. The word $w_i$ first obtains semantic information from its sense representations with the multi-dimensional attention,
\begin{equation}
\begin{split}
    \mathbf{q}^{l}_{i} = \mbox{MD-GAT}\left(\mathbf{h}_{i}^{l-1}, \ \left\{\mathbf{g}^{l}_{i,k} | s_{i,k} \in \mathcal{S}^{(w_i)} \right\} \right),
\end{split}
\end{equation}
and then updates its representation with a GRU:
\begin{equation}
    \mathbf{h}^l_{i} = \mbox{GRU}\left( \mathbf{h}_{i}^{l-1}, \ \mathbf{q}^l_{i}\right).
\end{equation}
The above GRU function and the GRU function in Eq. (\ref{eq:gru_sense}) have different parameters.

After multiple iterations, the final word representation $\mathbf{h}_{i}^L$ contains not only contextual word information but also semantic knowledge. For each sentence, we use $\mathbf{h}_{i}^a$ and $\mathbf{h}_{i}^b$ to denote the final word representation respectively.

\subsection{Sentence Matching Layer}

After obtaining the semantic knowledge enhanced word representation $\mathbf{h}_{i}^{a}$ and $\mathbf{h}_{i}^{b}$ for each sentence, we incorporate this word information into characters. Without loss of generality, we will use characters in sentence $\mathcal{C}^a$ to introduce the process. For each character $c_t^a$, we obtain $\mathbf{\hat{c}}^a_t$ by pooling the useful word information,

\begin{equation}
    \mathbf{\hat{c}}^a_t = \mbox{Att-Pooling}\left(\left\{ \mathbf{h}^{a}_{i}  |  w_i^a \in \mathcal{W}^{(c_t^a)} \right\} \right),
\end{equation}
where $\mathcal{W}^{(c_t^a)}$ is a set including words which contain the character $c_t^a$. The semantic knowledge enhanced character representation $\mathbf{y}_t$ is therefore obtained by 
\begin{equation}
    \mathbf{y}_t^a = \mbox{LayerNorm}\left( \mathbf{c}_t^a + \mathbf{\hat{c}}^a_t \right),
\end{equation}
where $\mbox{LayerNorm}(\cdot)$ denotes layer normalization, and $\mathbf{c}_t^a$ is the contextual character representation obtained using BERT described in Sec. \ref{sec:cwe}. 

For each character $c_t^a$, it aggregates information from sentence $\mathcal{C}^a$ and $\mathcal{C}^b$ respectively using multi-dimensional attention,
\begin{equation}
\begin{split}
    \mathbf{m}^{self}_{t} &= \mbox{MD-GAT} \left(\mathbf{y}_{t}^a, \ \{\mathbf{y}_{t'}^{a} | c_{t'}^a \in \mathcal{C}^a \} \right), \\
    \mathbf{m}^{cross}_{t} &= \mbox{MD-GAT} \ (\mathbf{y}_{t}^a, \ \{\mathbf{y}_{t'}^{b} | c_{t'}^b \in \mathcal{C}^b \} ) \ . \\
\end{split}
\end{equation}
The above multi-dimensional attention functions $\mbox{MD-GAT}(\cdot)$ share same parameters. With this sharing mechanism, the model has a nice property that, when the two sentences are perfectly matched, we have $\mathbf{m}_t^{self} \approx \mathbf{m}_t^{cross}$.

 We utilize the multi-perspective cosine distance \cite{wang2017bilateral} to compare $\mathbf{m}_t^{self}$ and $\mathbf{m}_t^{cross}$,
\begin{equation}
    d_k = \text{cosine}\left( \mathbf{w}_k^{cos} \odot \mathbf{m}_t^{self}, \mathbf{w}_k^{cos} \odot \mathbf{m}_t^{cross} \right),
\end{equation}
where $k \in \{1,2,\cdots,P\}$ ($P$ is number of perspectives).  $\mathbf{w}_k^{cos}$ is a parameter vector, which assigns different weights to different dimensions of messages. With $P$ distances $d_1, d_2, \cdots, d_P$, 
we can obtain the final character representation, 
\begin{equation}
   \mathbf{\hat{y}}_t^a = \text{FFN} \left(\left[\mathbf{m}_t^{self}, \mathbf{d}_t \right] \right),
\label{eq:ue}
\end{equation}
where $\mathbf{d}_t \triangleq [d_1, d_2, \cdots, d_P]$, and $\mbox{FFN}(\cdot) $ is a feed forward network with two layers.

Similarly, we can obtain the final character representation $\mathbf{\hat{y}}_t^b$ for each character $c_t^b$ in sentence $\mathcal{C}^b$.
Note that the final character representation contains three kinds of information: contextual information, word and sense knowledge, and character-level similarity. For each sentence $\mathcal{C}^a$ or $\mathcal{C}^b$, 
the sentence representation vector $\mathbf{r}^a$ or $\mathbf{r}^b$ is obtained using the attentive pooling of all final character representations for the sentence.

\subsection{Relation Classifier}

With two sentence vectors $\mathbf{r}^a$, $\mathbf{r}^b$, and the vector $\mathbf{c}^{\text{CLS}}$ obtained with BERT,  our model will predict the similarity of two sentences,
\begin{equation}
    p = \text{FFN} \left(\left[\mathbf{c}^{\text{CLS}}, \mathbf{r}^a, \mathbf{r}^b, \mathbf{r}^a \odot \mathbf{r}^b  , |\mathbf{r}^a - \mathbf{r}^b| \right]  \right),
\end{equation}
where $\mbox{FFN}(\cdot)$ is a feed forward network with two hidden layers and a sigmoid activation after output layer.

With $N$ training samples $\{\mathcal{C}^{a}_i, \mathcal{C}^{b}_i, y_i \}_{i=1}^N$, the training object is to minimize the binary cross-entropy loss,
\begin{equation}
    \mathcal{L} = - \sum_{i=1}^N \left( y_i \text{log}\left(p_i\right) + \left( 1 - y_i\right) \text{log}\left(1- p_i\right)  \right),
\end{equation}
where $y_i \in \{0,1\}$ is the label of the $i$-th training sample and $p_i \in[0,1]$ is the prediction of our model taking the sentence pair $\{\mathcal{C}^{a}_i, \mathcal{C}_i^{b}\}$ as input.





\section{Experiments}

\subsection{Experimental Setup}
\begin{table*}
\renewcommand\arraystretch{1.1} 
\centering{
\begin{tabular}{lcccccc}
\specialrule{0.1em}{1pt}{1pt}
\multirow{2}{*}{\textbf{Models}} & \multirow{2}{*}{\textbf{Pre-Training}}  & \multirow{2}{*}{\textbf{Interaction}} & \multicolumn{2}{c}{\textbf{BQ}} &  \multicolumn{2}{c}{\textbf{LCQMC}} \\

\cline{4-7}

 & & & \textbf{ACC.} & \textbf{F1} & \textbf{ACC.} & \textbf{F1} \\
\specialrule{0.1em}{1pt}{1pt}
Text-CNN\cite{he2016text} & $\times$ & $\times$ & 68.52  & 69.17 &  72.80 & 75.70 \\
BiLSTM\cite{mueller2016siamese} & $\times$ & $\times$ & 73.51 & 72.68 & 76.10 & 78.90 \\
Lattice-CNN \cite{lai2019lattice} & $\times$ & $\times$ & 78.20 & 78.30 & 82.14   & 82.41 \\

BiMPM \cite{wang2017bilateral} & $\times$ & $\surd$ & 81.85 & 81.73 & 83.30 & 84.90 \\
ESIM \cite{chen2017enhanced}  & $\times$ & $\surd$ & 81.93 & 81.87 & 82.58 & 84.49 \\
\textbf{LET} (Ours) & $\times$ & $\surd$ & \textbf{83.22} & \textbf{83.03} & \textbf{84.81} & \textbf{86.08} \\
\specialrule{0.1em}{1pt}{1pt}
BERT-wwm \cite{cui2019pre} & $\surd$ & $\surd$ & 84.89 & 84.29 & 86.80 & 87.78 \\
BERT-wwm-ext \cite{cui2019pre} & $\surd$ & $\surd$ & 84.71 & 83.94 & 86.68 & 87.71 \\
ERNIE \cite{sun2019ernie} & $\surd$ & $\surd$ & 84.67 & 84.20 & 87.04 & 88.06 \\\specialrule{0.0em}{0pt}{0pt}
BERT\cite{devlin2019bert} & $\surd$ & $\surd$ & 84.50 & 84.00 & 85.73 & 86.86 \\
\textbf{LET-BERT} (Ours) & $\surd$ & $\surd$ & \textbf{85.30}  & \textbf{84.98} & \textbf{88.38} & \textbf{88.85} \\
\specialrule{0.1em}{1pt}{1pt}
\end{tabular}
\caption{Performance of various models on LCQMC and BQ test datasets. The results are average scores using 5 different seeds. All the improvements over baselines are statistically significant ($p < 0.05$).}
\label{tab:main-res}
}
\end{table*}

\subsubsection{Dataset} We conduct experiments on two Chinese short text matching datasets: LCQMC \cite{liu2018lcqmc} and BQ \cite{chen2018bq}. 

LCQMC is a question matching corpus with large-scale open domain. It consists of 260068 Chinese sentence pairs including 238766 training samples, 8802 development samples and 12500 test samples. Each pair is associated with a binary label indicating whether two sentences have the same meaning or share the same intention. Positive samples are 30\% more than negative samples.  

BQ is a domain-specific large-scale corpus for bank question matching. It consists of 120000 Chinese sentence pairs including 100000 training samples, 10000 development samples and 10000 test samples. Each pair is also associated with a binary label indicating whether two sentences have the same meaning. The number of positive and negative samples are the same.



\subsubsection{Evaluation metrics} 
For each dataset, the accuracy (ACC.) and F1 score are used as the evaluation metrics. ACC. is the percentage of correctly classified examples. F1 score of matching is the harmonic mean of the precision and recall.

\subsubsection{Hyper-parameters}
The input word lattice graphs are produced by the combination of three segmentation tools: jieba~\cite{sun2012jieba}, pkuseg~\cite{pkuseg} and thulac~\cite{li2009punctuation}. We use the pre-trained sememe embedding provided by OpenHowNet~\cite{qi2019openhownet} with 200 dimensions. The number of graph updating steps/layers $L$ is 2 on both datasets, and the number of perspectives $P$ is 20. The dimensions of both word and sense representation are 128. The hidden size is also 128. The dropout rate for all hidden layers is 0.2. 
The model is trained by RMSProp with an initial learning rate of 0.0005 and a warmup rate of 0.1. The learning rate of BERT layer is multiplied by an additional factor of 0.1. As for batch size, we use 32 for LCQMC and 64 for BQ. \footnote{Our code is available at \url{https://github.com/lbe0613/LET}.}

\subsection{Main Results}

We compare our models with three types of baselines: representation-based models, interaction-based models and BERT-based models. The results are summarized in Table \ref{tab:main-res}. All the experiments in Table \ref{tab:main-res} and Table \ref{tab:ablation} are running five times using different seeds and we report the \textbf{average} scores to ensure the reliability of results. For the baselines, we run them ourselves using the parameters mentioned in \citet{cui2019pre}.

\textbf{Representation-based models} include three baselines Text-CNN, BiLSTM and Lattice-CNN. Text-CNN~\cite{he2016text} is one type of Siamese architecture with Convolutional Neural Networks (CNNs) used for encoding each sentence. BiLSTM~\cite{mueller2016siamese} is another type of Siamese architecture with Bi-directional Long Short Term Memory (BiLSTM) used for encoding each sentence. Lattice-CNN~\cite{lai2019lattice} is also proposed to deal with the potential issue of Chinese word segmentation. It takes word lattice as input and pooling mechanisms are utilized to merge the feature vectors produced by multiple CNN kernels over different $n$-gram contexts of each node in the lattice graph.


\textbf{Interaction-based models} include two baselines: BiMPM and ESIM. BiMPM~\cite{wang2017bilateral} is a bilateral multi-perspective matching model. It encodes each sentence with BiLSTM, and matches two sentences from multi-perspectives. BiMPM performs very well on some natural language inference (NLI) tasks. There are two BiLSTMs in ESIM~\cite{chen2017enhanced}. The first one is to encode sentences, and the other is to fuse the word alignment information between two sentences. ESIM achieves state-of-the-art results on various matching tasks. 
In order to be comparable with the above models, we also employ a model where BERT  in Fig. \ref{fig:net} is replaced by a traditional character-level transformer encoder, which is denoted as LET.

The results of the above models are shown in the first part of Table \ref{tab:main-res}. We can find that our model LET outperforms all baselines on both datasets. More specifically, the performance of LET is better than that of Lattice-CNN. Although they both utilize word lattices, Lattice-CNN only focuses on local information while our model can utilize global information. Besides, our model incorporates semantic messages between sentences, which significantly improves model performance. 
As for interaction-based models, although they both use the multi-perspective matching mechanism, LET outperforms BiMPM and ESIM. It shows the utilization of word lattice with our graph neural networks is powerful. 

\textbf{BERT-based models} include four baselines: BERT, BERT-wwm, BERT-wwm-ext and ERNIE. We compare them with our presented model LET-BERT. BERT is the official Chinese BERT model released by Google. BERT-wwm is a Chinese BERT with whole word masking mechanism used during pre-training. BERT-wwm-ext is a variant of BERT-wwm with more training data and training steps. ERNIE is designed to learn language representation enhanced by knowledge masking strategies, which include entity-level masking and phrase-level masking. LET-BERT is our proposed LET model where BERT is used as a character level encoder.

The results are shown in the second part of Table~\ref{tab:main-res}. We can find that the three variants of BERT (BERT-wwm, BERT-wwn-ext, ERNIE) all surpass the original BERT, which suggests using word level information during pre-training is important for Chinese matching tasks. Our model LET-BERT performs better than all these BERT-based models. Compared with the baseline BERT which has the same initialization parameters, the ACC. of LET-BERT on BQ and LCQMC is increased by 0.8\% and 2.65\%, respectively.  It shows that utilizing sense information during fine-tuning phrases with LET is an effective way to boost the performance of BERT for Chinese semantic matching.

We also compare results with K-BERT~\cite{liu2020k}, which regards information in HowNet as triples \{word, contain, sememes\} to enhance BERT, introducing soft position and visible matrix during the fine-tuning and inferring phases. The reported ACC. for the LCQMC test set of K-BERT is 86.9\%. Our LET-BERT is 1.48\% better than that. Different from K-BERT, we focus on fusing useful information between word and sense.

\subsection{Analysis}

\begin{table}[t]

\centering
\setlength{\tabcolsep}{5mm}{\begin{tabular}{lccc}
\specialrule{0.1em}{1pt}{1pt}
\textbf{Seg.}    & \textbf{Sense}    & \textbf{ACC.} & \textbf{F1} \\
\specialrule{0.1em}{1pt}{1pt}
jieba   & $\surd$  &  87.84    &  88.47  \\
pkuseg  & $\surd$  &  87.72    &  88.40  \\
thulac  & $\surd$  &  87.50    &  88.27  \\
lattice & $\surd$  & \textbf{88.38}    &  \textbf{88.85} \\
lattice & $\times$ &  87.68    &  88.40  \\
\specialrule{0.1em}{1pt}{1pt}
\end{tabular}
\caption{Performance of using different segmentation on LCQMC test dataset.}
\label{tab:ablation}}
\end{table}

In our view, both multi-granularity information and semantic information are important for LET. If the segmentation does not contain the correct word, our semantic information will not exert the most significant advantage.

Firstly, to explore the impact of using different segmentation inputs, we carry out experiments using LET-BERT on LCQMC test set. As shown in Table \ref{tab:ablation}, when incorporating sense information, improvement can be observed between lattice-based model (the fourth row) and word-based models: jieba, pkuseg and thulac. The improvements of lattice with sense over other models in Table \ref{tab:ablation} are all statistically significant ($p < 0.05$). The possible reason is that lattice-based models can reduce word segmentation errors to make predictions more accurate.

Secondly, we design an experiment to demonstrate the effectiveness of incorporating HowNet to express the semantic information of words. In the comparative model without HowNet knowledge, the sense updating module in SaGT is removed, and we update word representation only by a multi-dimensional self-attention. The last two rows in Table \ref{tab:ablation} list the results of combined segmentation (lattice) with and without sense information. The performance of integrating sense information is better than using only word representation. More specifically, the average absolute improvement in ACC. and F1 scores are 0.7\% and 0.45\%, respectively, which indicates that LET has the ability to obtain semantic information from HowNet to improve the model's performance. 
Besides, compared with using a single word segmentation tool, semantic information performs better on lattice-based model. The probable reason is lattice-based model incorporates more possible words so that it can perceive more meanings.

We also study the role of GRU in SaGT. The ACC. of removing GRU in lattice-based model is 87.82\% on average, demonstrating that GRU can control historical messages and combine them with current information. Through experiments, we find that the model with 2 layers of SaGT achieves the best. It indicates multiple information fusion will refine the message and make the model more robust.

\subsubsection{Influences of text length on performance}

As listed in Table \ref{tab:length}, we can observe that text length also has great impacts on text matching prediction. The experimental results show that the shorter the text length, the more obvious the improvement effect of utilizing sense information. The reason is, on the one hand, concise texts usually have rare contextual information, which is difficult for model to understand. However, HowNet brings a lot of useful external information to these weak-context short texts. Therefore, it is easier to perceive the similarity between texts and gain great improvement.
 On the other hand, longer texts may contain more wrong words caused by insufficient segmentation, leading to incorrect sense information. Too much incorrect sense information may confuse the model and make it unable to obtain the original semantics.
 
\begin{table}[t]
\centering
\resizebox{\linewidth}{!}{
\begin{tabular}{ccccc}
\specialrule{0.1em}{1pt}{1pt}
\multirow{2}{*}{\textbf{text length}} & \multirow{2}{*}{\textbf{\begin{tabular}[c]{@{}c@{}}number of \\ samples\end{tabular}}} & \multicolumn{2}{c}{\textbf{ACC.}}&\multirow{2}{*}{\textbf{RER(\%)}} \\ \cline{3-4} 
                           &                                           & w/o sense     & sense  &   \\ \specialrule{0.1em}{1pt}{1pt}
     $<16$                        &        2793                          &       88.99                        &    90.05     &\textbf{9.63} \\\specialrule{0em}{1pt}{1pt}
          $16-18$                    &        3035                        &    88.49                       &     89.25    &6.60 \\\specialrule{0em}{1pt}{1pt}
           $19-22$                   &      3667                              &       88.58                 &    89.04    &4.03 \\\specialrule{0em}{1pt}{1pt}
                
        $>22$                   &    3005  &     84.53        &       85.13 &3.88\\\specialrule{0.1em}{1pt}{1pt}
\end{tabular}
}
\caption{Influences of text length on LCQMC test dataset. Relative error reduction (RER) is calculated by $\frac{\text{sense} - \text{w/o sense}}{100 - \text{w/o sense}} \times 100\%$.}
\label{tab:length}
\end{table}

\subsubsection{Case study}
\begin{figure}[h]
    \centering
    \includegraphics[width=1\columnwidth]{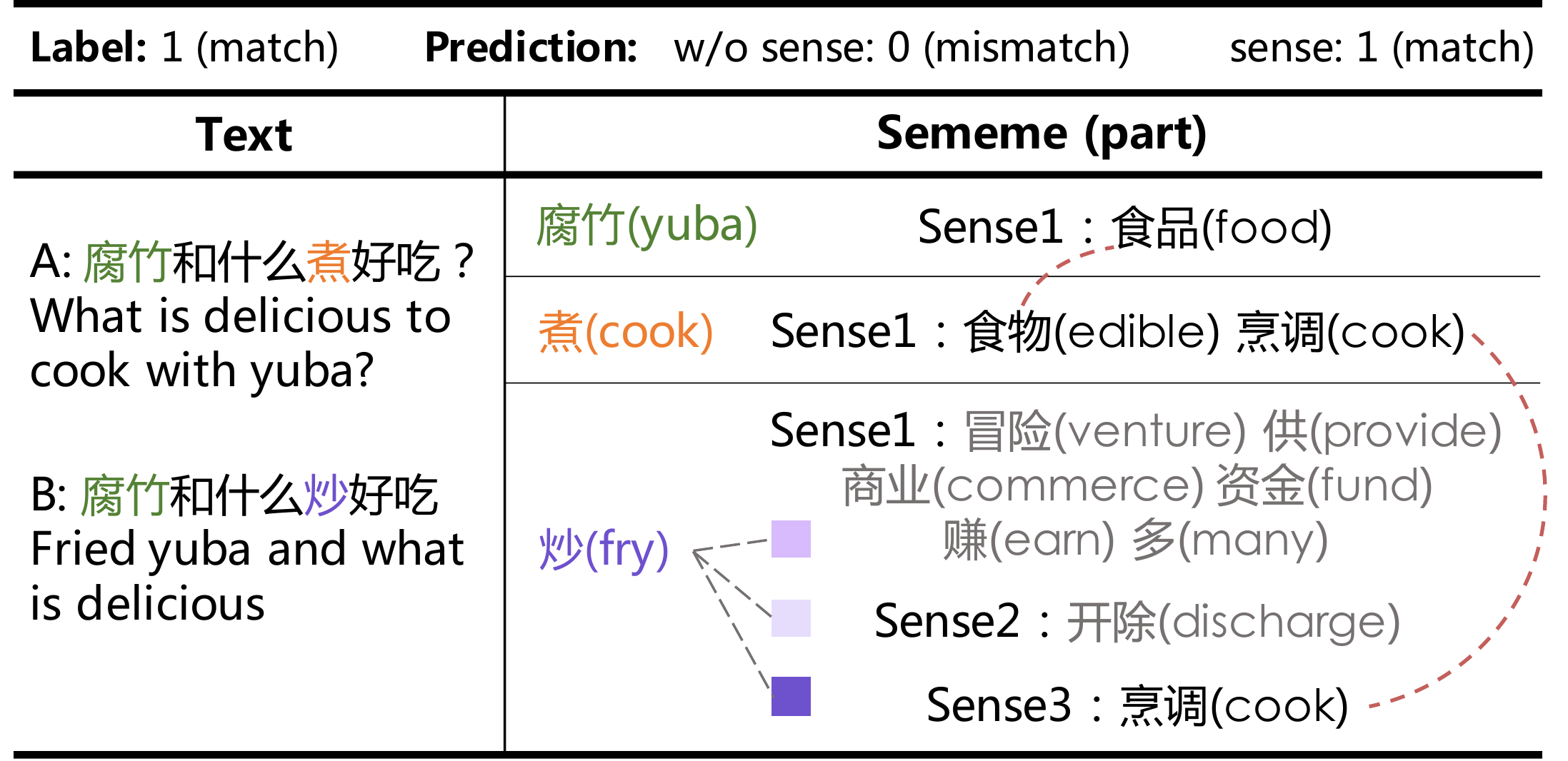}
    \caption{An example of using sense information to get the correct answer.}
    \label{fig:case}
 
\end{figure}

We compare LET-BERT between the model with and without sense information (see Fig.~\ref{fig:case}). The model without sense fails to judge the relationship between sentences which actually have the same intention, but LET-BERT performs well. We observe that both sentences contain the word  ``yuba'', which has only one sense described by sememe {\tt food}. Also, the sense of ``cook'' has a similar sememe {\tt edible} narrowing the distance between texts. Moreover, the third sense of ``fry'' shares the same sememe {\tt cook} with the word ``cook''. It provides a powerful message that makes ``fry'' attend more to the third sense.

\section{Conclusion}
In this work, we proposed a novel linguistic knowledge enhanced graph transformer for Chinese short text matching. Our model takes two word lattice graphs as input and integrates sense information from HowNet to moderate word ambiguity. The proposed method is evaluated on two Chinese benchmark datasets and obtains the best performance. The ablation studies also demonstrate that both semantic information and multi-granularity information are important for text matching modeling.

\section{Acknowledgments}
We thank the anonymous reviewers for their thoughtful comments.  This work has been supported by No. SKLMCPTS2020003 Project.

\bibliography{reference}

\end{document}